# Fine-Tuning BERT for Automatic ADME Semantic Labeling in FDA Drug Labeling to Enhance Product-Specific Guidance Assessment

Yiwen Shi[1], Jing Wang[2], Ping Ren[2], Taha ValizadehAslani[3], Yi Zhang[2], Meng Hu[2], Hualou Liang[4*]

[1] College of Computing and Informatics, Drexel University, Philadelphia, PA, United States
[2] Office of Research and Standards, Office of Generic Drugs, Center for Drug Evaluation and Research, United States Food and Drug Administration, Silver Spring, MD, United States
[3] Department of Electrical and Computer Engineering, College of Engineering, Drexel University, Philadelphia, PA, United States
[4] School of Biomedical Engineering, Science and Health Systems, Drexel University, Philadelphia, PA, United States

*Corresponding author: hualou.liang@drexel.edu

## Abstract

Product-specific guidances (PSGs) recommended by the United States Food and Drug Administration (FDA) are instrumental to promote and guide generic drug product development. To assess a PSG, the FDA assessor needs to take extensive time and effort to manually retrieve supportive drug information of absorption, distribution, metabolism, and excretion (ADME) from the reference listed drug labeling. In this work, we leveraged the state-of-the-art pre-trained language models to automatically label the ADME paragraphs in the pharmacokinetics section from the FDA-approved drug labeling to facilitate PSG assessment. We applied a transfer learning approach by fine-tuning the pre-trained Bidirectional Encoder Representations from Transformers (BERT) model to develop a novel application of ADME semantic labeling, which can automatically retrieve ADME paragraphs from drug labeling instead of manual work. We demonstrated that fine-tuning the pre-trained BERT model can outperform the conventional machine learning techniques, achieving up to 11.6% absolute F1 improvement. To our knowledge, we were the first to successfully apply BERT to solve the ADME semantic labeling task. We further assessed the relative contribution of pre-training and fine-tuning to the overall performance of the BERT model in the ADME semantic labeling task using a series of analysis methods such as attention similarity and layer-based ablations. Our analysis revealed that the information learned via fine-tuning is focused on task-specific knowledge in the top layers of the BERT, whereas the benefit from the pre-trained BERT model is from the bottom layers.

# 1. Introduction

The U.S. Food and Drug Administration (FDA) publishes Product-Specific Guidance[1] (PSG) to reflect the FDA's current thinking and expectations on how to establish bioequivalence between a test product and the corresponding reference listed drug. The PSGs can facilitate generic drug product development, Abbreviated New Drug Application (ANDA) submission and approval, and ultimately provide access to safe, effective and affordable generic drugs to the public. To enhance the PSG assessment process for generic drug products, it is desirable to identify and automate the labor-intensive work to collect supportive information from drug labeling during the PSG assessment (Shi et al., 2021). For example, retrieving information from drug labeling, such as absorption, distribution, metabolism, and excretion (ADME) of Reference Listed Drug (RLD) products is an important step in the PSG assessment.

**Why do we need automatic ADME semantic labeling?**

Semantic labeling involves assigning class labels to sentences or paragraphs based upon their semantic meaning (Hulsebos et al., 2019; Pham et al., 2016; Ruemmele et al., 2018; Trabelsi et al., 2020). ADME semantic labeling is indispensable in PSG assessment that can guide generic product development by promoting timely approval of ANDA submission and drug price competition. Therefore, it is an important task to enable further generic drug development via facilitating increased drug price competition in the market for prescription drugs through the approval of lower-cost and high-quality generic drugs relative to the brand-name drugs.

As of this writing, there are nearly 5,000 brand-name products[2]; only 1,201 drug labels have the ADME subsection titles (for example, see Figure A.1); many do not have explicit ADME labeling (Figure 1). Therefore, the scarcity of the ADME labeling in FDA drugs calls for the development of automatic ADME semantic labeling. Currently, the ADME information is exclusively manually collected from drug labeling by the PSG assessor, which is time-consuming and labor-intensive. To meet the GDUFA commitment to PSG assessment[3], development of a data analytics tool that can identify and automate the ADME information collection/retrieval from drug labeling during the PSG assessment will significantly facilitate the fulfillment of the FDA's GDUFA commitment and the FDA's aim to support in generic drug development. To our knowledge, we are the first to successfully apply BERT to solve the ADME semantic labeling task.

**Why is automatic ADME semantic labeling challenging?**

DailyMed[4] is a public data source for FDA-approved drug labeling, which contains a broad array of information for the study of drug products and safety information, such as indications (Fung et al., 2013), boxed warning, and adverse reactions (Bisgin et al., 2011). It provides electronic drug labeling following the Structured Product Labeling (SPL) standard, which specifies various drug label sections by Logical Observation Identifiers Names and Codes (LOINC)[5]. LOINC code has been widely used to extract information from drug labeling but can only be used for high-level sections (e.g., Indications and Usage, Dosage and Administration, Pharmacokinetics), as the content in each section is free narrative text. Since the ADME doesn't have its own LOINC, we can only extract its parent section, the pharmacokinetics section, by LOINC code ("43682-4"). However, the pharmacokinetics section contains not only the ADME related information, but also other topics such as drug-drug interactions, specific content for a certain age, gender, health condition (hepatic/renal impairment), and so on, which need to be excluded. Such ADME irrelevant information brings additional difficulties to automatically labeling ADME paragraphs in the pharmacokinetics section.

---

[1] https://www.accessdata.fda.gov/scripts/cder/psg/index.cfm
[2] https://www.fda.gov/drugs/guidances-drugs/product-specific-guidances-generic-drug-development
[3] https://www.fda.gov/media/101052/download
[4] https://dailymed.nlm.nih.gov/dailymed
[5] https://www.fda.gov/industry/structured-product-labeling-resources/section-headings-loinc

Moreover, the writing style of drug labeling is rather diverse. Some have explicit subsection titles within the pharmacokinetics section (Figure A.1), but many do not (see an example shown in Figure 1). As a result, the rule-based method by simply searching for keywords in the paragraphs for labeling does not work well. For example, the sentence of "*The drug passes into the aqueous humor of the eye achieving a concentration of approximately one tenth of plasma concentrations.*" is about "Distribution", yet does not contain the keyword "Distribution". In another example, the passage of "*Because of the absorption-rate limited kinetics of insulin mixtures, a true half-life cannot be accurately estimated from the terminal slope of the concentration versus time curve.*" does include the keyword "Absorption", but it is about "Excretion". These two examples underscore the difficulty with the rule-based method as the ADME labeling is determined by the meaning of the passage, rather than the simple keywords. Furthermore, conventional machine learning algorithms, such as Logistic Regression, Linear Support Vector Classification (SVC), and Random Forest, which are commonly used for text classification, are also not adequate for the ADME labeling, as shown in the Results section, indicating that it is indeed a hard classification task.

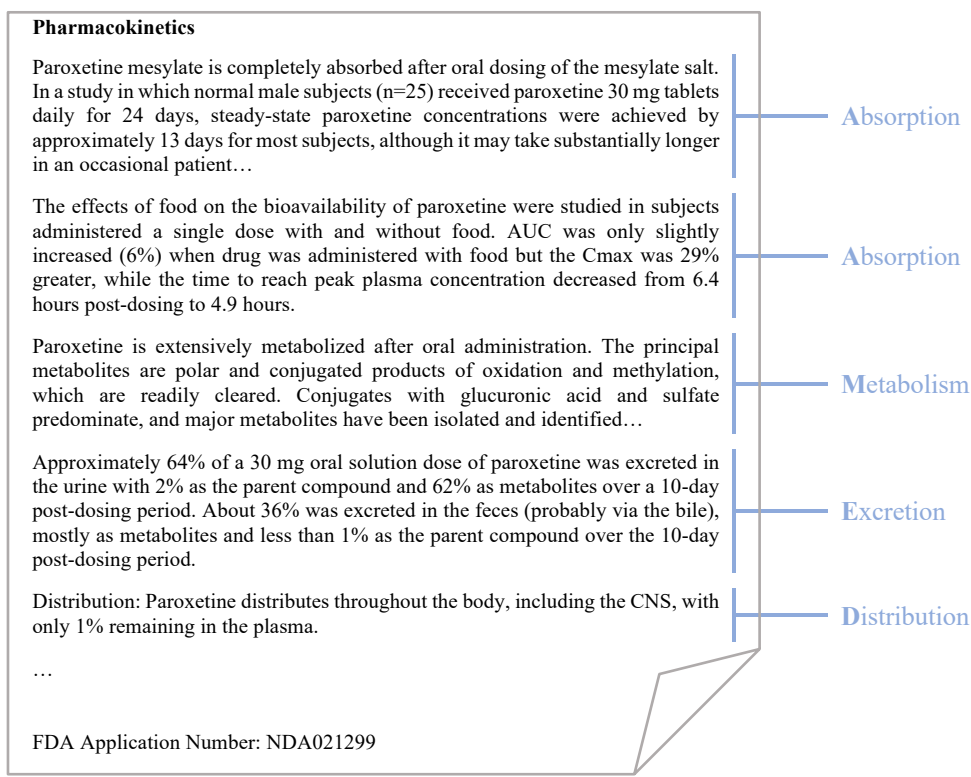

Figure 1: Example of pharmacokinetics section in drug labeling without ADME title

As a result, we proposed text classification by transfer learning to automatically label ADME paragraphs with their semantic meaning. Transfer learning is a powerful approach in Natural Language Processing (NLP), which leverages prior knowledge with a pre-trained model into a new task. Pre-trained language models have transformed the landscape of NLP (Brown et al., 2020; Devlin et al., 2019; Raffel et al., 2020, *inter alia*). Large pre-trained models, such as BERT (Devlin et al., 2019), are a breakthrough in the use of deep learning in NLP, which brought remarkable advancement to many NLP tasks, including semantic analysis. The pre-trained model uses unsupervised training on a large corpus of unlabeled text to learn about the structure of language, such as the basic semantic and syntax information (Clark et al., 2019). A common approach to adopt it in downstream tasks is further training a pre-trained model on a relatively smaller dataset, and this process is known as fine-tuning. Fine-tuning usually reuses the pre-trained model's parameters as a starting point by plugging in the task-specific input/output and adding one task-specific layer to train with supervised data. In this work, we adopted the fine-tuning approach in our task and made the following contributions:

- We developed a novel application to automatically label ADME paragraphs in the pharmacokinetics section from drug labeling to accelerate Product-Specific Guidance assessment.
- We used a transfer learning approach by fine-tuning the BERT model to learn the ADME semantic labeling, which obtained a better performance than traditional machine learning models (e.g., with F1 0.9268 for BERT and 0. 8110 for Random Forest).
- We assessed the relative contribution of pre-training and fine-tuning to the ADME semantic labeling task to provide an in-depth understanding of the model's components. We found that both procedures contributed to overall performance, yet the information learned by BERT's fine-tuning is focused on task-specific knowledge in the top layers, whereas the benefit from the pre-trained BERT model is from the bottom layers.

## 2. Related Work

**ADME Prediction:** Successful drug discovery requires control and optimization of compound properties related to pharmacokinetics, pharmacodynamics, and safety. Optimizing ADME properties is an integral process in drug discovery and optimization, in which deep learning plays an important role in drug discovery. Especially, when the experimental capacity and the prediction of in vitro ADME assays is often limited, in silico ADME models can overcome this shortcoming (Danielson et al., 2017; Gleeson et al., 2011; Göller et al., 2020). The latest generation of deep learning methods such as deep neural networks (DNNs) can further improve the quality of the models and decision making. Wenzel et al. (2019) proposed a multitask DNN model in the field of industrial ADME-Tox predictions, which performed statistically superior on most studied data sets in comparison to single-task models. Zhou et al. (2019) analyzed the impact of different hyperparameter settings on the generalization of DNN models for predicting quantitative structure−activity relationships (QSARs) and molecular activities, including ADME properties. To the best of our knowledge, there is no study analyzing ADME from a natural language semantic analysis perspective.

**Transfer Learning in Semantic Analysis:** Semantic analysis is an important task in NLP. It solves the computational processing of opinions, emotions, and subjectivity. Transfer learning showed empirical improvements in semantic analysis from pre-trained word embedding trained on a large corpus. Word2Vec (Mikolov et al., 2013) predicts the current word based on a continuously distributed representation of the context. GloVe (Pennington et al., 2014) is an approach to combining both the global statistics of matrix factorization with local context-based learning. Bojanowski et al. (2017) represented a word as a bag of character n-grams helped the rare words and out-of-vocabulary (OOV) words obtained better vector representations. However, the word embeddings generated from the language models above are context-free, which means they cannot handle homonyms, a word has different meanings in a different context. Moreover, the trivial way to build a paragraph vector for semantic analysis is taking the sum or the average of word embeddings, but it loses the words' position information.

More recently, the transformer-based language models became a breakthrough in the use of deep learning in NLP, which brought remarkable advancement to many downstream tasks. OpenAI GPT (Radford et al., 2019), and BERT (Devlin et al., 2019), have shown their effectiveness in using contextualized word representation. Especially, BERT has achieved state-of-the-art results in many NLP benchmarks, including semantic analysis benchmark SST-2 (Socher et al., 2013). Prior works applied BERT in semantic analysis of the biomedical domain as well (Peng et al., 2019), such as semantic classification for drug reviews (Biseda & Mo, 2020; Colón-Ruiz & Segura-Bedmar, 2020), adverse drug reactions detection (Breden & Moore, 2020; Fan et al., 2020; Hussain et al., 2021), biomedical named entity recognition (Khan et al., 2020), which showed BERT models can achieve competitive performance across all different tasks and datasets in the biomedical domain. Our study focused on classifying ADME information from drug labeling by its semantic meaning, which provided a novel application to retrieving ADME information from FDA-approved drug labeling by fine-tuning pre-trained BERT.

## 3. Methods

### 3.1 Dataset

DailyMed is a free drug information resource provided by the U.S. National Library of Medicine (NLM) that consists of digitized versions of drug labeling as submitted to the FDA. It provides access to electronic drug labeling via RESTful API. The electronic drug labeling follows the SPL standard, which specifies various drug label sections by LOINC. In this study, we focused on the pharmacokinetics section (LOINC code: 43682-4) of the drug labeling which contains ADME content as free narrative text. We broke down the content into the smallest text segments identifiable by the XML tags (e.g., <title>, <paragraph> and <item>). Each segment ended with '.' was identified as a paragraph.

To prepare for the training data for model, we used regular expressions to detect two types of ADME subsection titles (shown in Figure A.1) in the pharmacokinetics section from 1,201 drug labeling, which resulted in 5,777 ADME paragraphs and 5,232 paragraphs under "Other" topics (e.g., "specific populations", "drug interaction studies", etc.). Table 1 (unshaded) showed a breakdown of the number of samples in each class of the dataset. Further details about drug labeling selection and the annotation methods are provided in the Data Preprocessing section.

Table 1: Summary statistics of the datasets. Shaded are the manually-annotated unseen dataset.

| Topic | Count | |
|---|---|---|
| Absorption | 1955 | 190 |
| Distribution | 1213 | 100 |
| Metabolism | 1137 | 104 |
| Excretion | 1472 | 138 |
| Other | 5232 | 620 |

To evaluate the model generalization ability to unseen data, we additionally created a manually-annotated dataset by randomly sampling unlabeled drug labels (see Figure 1 for example) and assigning ADME labels by hand, which resulted in 1,152 paragraphs that the ADME title cannot be simply detected by regular expression (Table 1, shaded). This unseen dataset, which imitates the task described in Figure 1, provides an independent validation of the model performance.

### 3.2 Data Preprocessing

**Drug labeling selection**: A workflow for drug labeling selection is illustrated in Figure 2. Overall, there are three major steps to select the drug labels for training the model.

1. The drug labeling should belong to FDA approved drugs with New Drug Application (NDA) number. The drug label without an NDA FDA Application Number was removed. We used the FDA Application Number as the identifier, and only kept the latest version of drug labeling if multiple versions exist.

2. The pharmacokinetics section was extracted with LOINC Code "43682-4". Drug labeling was removed if there was no content in Pharmacokinetics Section.

3. The regular expressions were used to detect the ADME section titles in pharmacokinetics content. If no ADME section title was found, the drug labeling was removed.

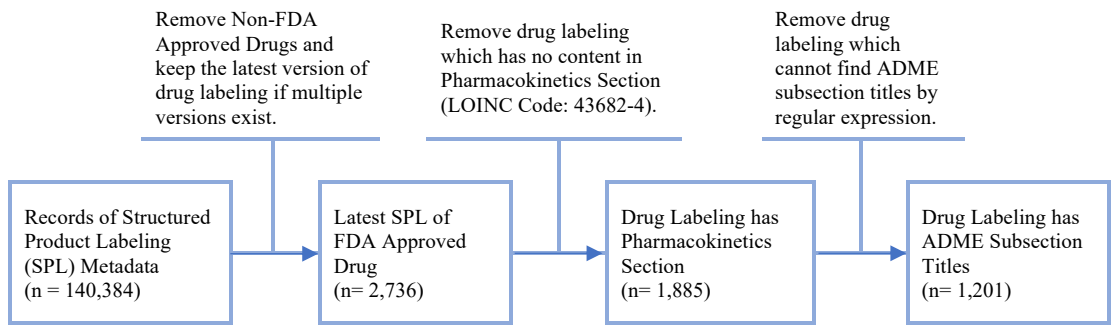

Figure 2: Workflow for Drug Labeling Selection from DailyMed. (Metadata is downloaded on August 18, 2021)

**ADME paragraphs annotation:** There are two types of ADME subsection titles that can be detected by regular expressions (Figure A.1). One type of title is outside paragraphs, in which case, we detected the section title with the regular expression

"^(absorption|distribution|metabolism|excretion|…)$", and annotated the paragraphs before the next section title as the title they follow. For example (Figure A.1A), the paragraph "*In pediatric patients with ALL, oral absorption of methotrexate appears to be dose dependent…*" follows the section title 'Absorption' and before "Distribution", which is annotated as "Absorption". Another type of title is inside paragraphs, in which case, we detected the section title with the regular expression "^(absorption|distribution|metabolism|excretion|…)\s*(:|-)", and annotated the paragraph with the title at the beginning. For example (Figure A.1B), the paragraph "*The estimated oral bioavailability of immediate release metoprolol is about 50% because of pre-systemic metabolism…*" is annotated as "Absorption". Note that we considered "Elimination" as "Excretion" and "Food Effect" as "Absorption". Paragraphs under titles other than ADME (e.g., "specific populations", "drug interaction studies", etc.) were annotated as "Other".

### 3.3 Models

**Rule-based Method:** The rule-based method encoded human knowledge in if-then statements for specific rules. For the ADME labeling task, we used the rule-based method as a baseline, which used a simple logic by searching keywords in the paragraphs to identify the ADME topic.

**Machine Learning Models:** Conventional machine learning algorithms, such as Logistic Regression, Linear Support Vector Classification (SVC), and Random Forest with the Term Frequency-Inverse Document Frequency (TF-IDF) (Chowdhury, 2010) representation, were also used for comparison. Note that TF-IDF is a statistical measure based on the concurrence of terms, which fails to identify syntactic and semantic relationships between words in the documents.

**BERT:** BERT (Devlin et al., 2019) is a deep neural network that uses the transformer encoder architecture (Vaswani et al., 2017) to learn contextualized embeddings of the input text. BERT utilizes WordPiece (Wu et al., 2016) for tokenization to convert the words in the input text to tokens that can be understood by the model. WordPiece is a subword-based tokenization that can cover a wider range of OOV words (Sennrich et al., 2016). BERT is a stack of transformer architecture, and the transformer is entirely based on attention. An attention function is a mapping from a query and a set of key-value pairs to an output. The output is computed as a weighted sum of the values, where the weight on values are dot products of the query with all keys, divide by $\sqrt{d_k}$, ($d_k$: dimension of the key). A softmax is applied to map the weights to [0,1] and to ensure they sum to 1 over the whole sequence. Attention weights can be viewed as how important every other token is. It suggests that BERT's attention maps have a fairly thorough representation of syntactic information. In each transformer encoder layer, the self-attention mechanism learns the relationship between different token representations. Instead of having a single attention head, in BERT, multiple attention heads exist and the output of them is concatenated and linearly projected. This is called multi-head attention. The multi-head attention mechanism allows the model to jointly attend to information from different representation subspaces at different positions. There are two steps in training BERT: pre-training and fine-tuning. During the pre-training stage, BERT learns the distribution of the words in the general language from the large corpus of unlabeled text including BooksCorpus (800M words) and English Wikipedia (2,500M words) using masked language modeling and next-sentence prediction methods. At the fine-tuning stage, BERT is initialized with the pre-trained weights from the previous stage, and all the parameters are fine-tuned by labeled data from downstream tasks. There are two architectures available for BERT: BERT-base and BERT-large. The difference between BERT-base and BERT-large is in the number of encoder layers, attention heads, and parameters. BERT-base model has 12 encoder layers, 12 attention heads, 768 hidden sizes, and 110 million parameters, whereas BERT-large has 24 layers of encoders, 16 attention heads, 1024 hidden sizes, and 340 million parameters.

In this paper, we used the forementioned approach to fine-tune the BERT model and compared its performance with the rule-based method and three traditional machine learning algorithms: Logistic Regression, Linear SVC, and Random Forest. Moreover, we used the BERT-base-uncased[6] to study the relative contribution of pre-training and fine-tuning by analyzing the attention of the BERT model and performing a series of ablation studies.

### 3.4 Evaluation

To assess the performance of different models for ADME semantic labeling, we performed a stratified 5-fold cross-validation where the relative class frequencies are approximately preserved in each fold. Specifically, we split the dataset into five folds, each contains approximately the same percentage of samples of each target class as the complete set. We used four folds for training and validation (three-fold for model training, and one-fold for validation to tune hyper-parameters), and the remaining for testing in each run for five independent runs. In addition, we evaluated the models with an unseen, independent testing dataset which was manually collected and annotated. We used macro-averaged precision, recall, and F1 score as the performance metrics. All the results we reported were the average of the five independent runs, together with its standard deviation (STD). All experiments were run on a single Nvidia Tesla P100-PCIE-16GB.

## 4. Results

In this section, we presented the experiments conducted for learning the ADME semantic labeling with two baselines: rule-based method and conventional machine learning models, as well as fine-tuning the pre-trained BERT models (Section 4.1). We then demonstrated the relative contribution to the overall performance of the BERT models by each component: fine-tuning (Section 4.2) and pre-training (Section 4.3).

### 4.1 ADME Semantic Labeling

**Rule-based Method (Baseline 1):** Each ADME topic has some keywords that can be used to design a rule-based method for ADME labeling (shown in Table 2). If a paragraph matched any keyword, we labeled it as ADME accordingly. If keywords of multiple ADME topics were found in one paragraph, we randomly selected one to label it. If no keyword was detected, we labeled the paragraph as "Other". We used the rule-based method to provide a baseline comparison with the conventional machine learning methods and the BERT models.

**Machine Learning Models (Baseline 2):** As a comparison, we performed three traditional machine learning algorithms: Logistic Regression, Linear SVC, and Random Forest, with the word embedding method TF-IDF. We tuned the hyperparameters of each model and set the TF-IDF feature's length as 128 which is the same as the maximum sequence length in the BERT model.

---

[6] https://huggingface.co/bert-base-uncased

Table 2: Keywords of ADME topic used for the rule-based method

| Topic | Keywords |
|---|---|
| Absorption | absorption, absorb, food |
| Distribution | distribution, distribute |
| Metabolism | metabolism, metabolize |
| Excretion | excretion, elimination, excrete, eliminate |

**BERT:** We used BERT-base and BERT-large models to fine-tune the ADME semantic labeling task. The models were pre-trained and initialized with the parameters released by Devlin et al. (2019), which can be accessed from Huggingface (Wolf et al., 2020). We used configurations mostly consistent with the recommendations in the original release. We grid searched a batch size of {10, 16, 32, 64}, and a learning rate of {5e-6, 1e-5, 3e-5, 5e-5}. In the pharmacokinetics section, as shown in Figure 3, 95% of the paragraph length is less than 128, so we set the maximum sequence length as 128.

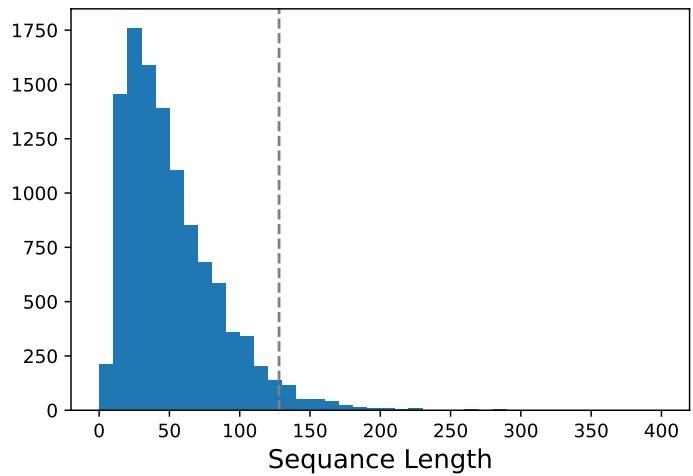

Figure 3: Sequence Length of Paragraph in Pharmacokinetics Section

**Results:** Table 3 showed that BERT models (with F1 0.9134 for BERT-base and 0.9268 for BERT-large) outperformed both the rule-based method (with F1 0.6588) and the traditional machine learning models (with F1 0.8053, 0.7969, and 0.8110 for Logistic Regression, Linear SVC, and Random Forest, respectively). Among the conventional models, Random Forest had the best performance. Moreover, the BERT-base model had the lowest standard deviation of performance metrics, which showed better stability.

To evaluate the model generalization ability on an unseen dataset, we tested the models with a manually collected and annotated dataset containing 1,152 paragraphs that the ADME title cannot be detected by regular expression. We observed that the BERT models (with F1 0.9024 for BERT-base and 0.9161 for BERT-large) still performed better than both the rule-based method (with F1 0.6766) and the traditional machine learning models (with F1 0.7935, 0.7899, and 0.7953 for Logistic Regression, Linear SVC, and Random Forest, respectively). We also revisited the drug labeling without the ADME title shown in Figure 1. Table 4 showed the BERT model labeled all five paragraphs correctly, but Random Forest missed one.

Table 3: Comparison of performance metrics (STD) with different models. Shaded are the results of the unseen testing dataset.

| Model | Precision | Recall | F1 |
|---|---|---|---|
| Rule-based | 0.7281 (0.0076) | 0.6337 (0.0121) | 0.6588 (0.0106) |
| Logistic Regression | 0.8449 (0.0094) | 0.7757 (0.0089) | 0.8053 (0.0081) |
| Linear SVC | 0.8382 (0.0073) | 0.7686 (0.0077) | 0.7969 (0.0067) |
| Random Forest | 0.8435 (0.0100) | 0.7867 (0.0111) | 0.8110 (0.0103) |
| BERT-base | 0.9118 (0.0047) | 0.9155 (0.0068) | 0.9134 (0.0029) |
| BERT-large | **0.9284 (0.0100)** | **0.9259 (0.0046)** | **0.9268 (0.0055)** |
| Rule-based | 0.7134 (-) | 0.6665 (-) | 0.6766 (-) |
| Logistic Regression | 0.8254 (0.0073) | 0.7685 (0.0125) | 0.7935 (0.0105) |
| Linear SVC | 0.8262 (0.0046) | 0.7641 (0.0090) | 0.7899 (0.0076) |
| Random Forest | 0.8161 (0.0084) | 0.7788 (0.0103) | 0.7953 (0.0091) |
| BERT-base | 0.8893 (0.0029) | 0.9168 (0.0026) | 0.9024 (0.0015) |
| BERT-large | **0.9051 (0.0036)** | **0.9283 (0.0070)** | **0.9161 (0.0046)** |

Table 4: Example of ADME semantic labeling prediction with different models

| Paragraph | True Label | Prediction | |
|---|---|---|---|
| | | Random Forest | BERT |
| Paroxetine mesylate is completely absorbed after oral dosing of the mesylate salt. In a study in which normal male subjects (n=25) … | Absorption | Other | Absorption |
| The effects of food on the bioavailability of paroxetine were studied in subjects administered a single dose with and without food… | Absorption | Absorption | Absorption |
| Paroxetine is extensively metabolized after oral administration. The principal metabolites are polar and conjugated products of oxidation … | Metabolism | Metabolism | Metabolism |
| Approximately 64% of a 30 mg oral solution dose of paroxetine was excreted in the urine with 2% as the parent compound and 62% as metabolites… | Excretion | Excretion | Excretion |
| Distribution: Paroxetine distributes throughout the body, including the CNS, with only 1% remaining in the plasma. | Distribution | Distribution | Distribution |

**Learning Curve**: We further substantiated that the BERT model outperformed conventional machine learning methods by comparing the learning curves of the Random Forest model with the BERT-base models. We randomly selected 200 records from each class and set them aside as the testing dataset, then observed the performance when the training data per class gradually increased.

**Results**: Figure 4 showed the performance improved when training data size increased for both BERT-base and Random Forest. BERT-base outperformed Random Forest with different sizes of the training dataset. Overall, the results showed that fine-tuning BERT model performed better than the conventional machine learning techniques for ADME semantic labeling on drug labeling datasets.

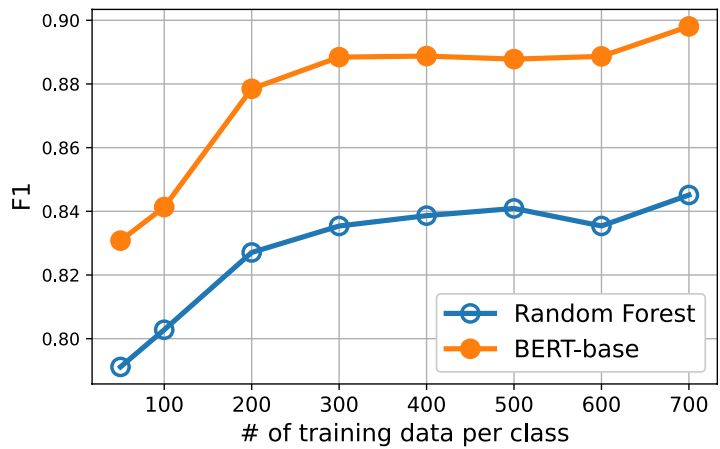

Figure 4: Learning curve of test performance with increasing amounts of training data

### 4.2 Impact of Fine-tuning

Fine-tuning has a significant effect on the performance of the ADME semantic labeling task. Table 5 showed that the fine-tuned BERT (with F1 0.9134) outperformed pre-trained BERT (with F1 0.1392) by a significant margin. We designed a series of experiments to find out how fine-tuning impacted the model by evaluating the attentions difference between the pre-trained and fine-tuned BERTs. We also performed a layer-wise ablation study to quantify the impact of fine-tuning in different layers. We used the BERT-base model in these experiments.

Table 5: F1 score of BERT-base model with different initialization

| Pre-trained BERT | Fine-tuned, initialized with | | |
| --- | --- | --- | --- |
|  | Uniform | Truncated Normal | Pre-trained BERT |
| 0.1392 | 0.7450 | 0.8522 | 0.9134 |

**Attentional changes**: To compare the attentions between the pre-trained BERT model and the fine-tuned one, we computed the cosine similarity between flattened attentions. The attentions which have fewer changes get higher cosine similarity, whereas the attentions that have more changes get lower cosine similarity. To observe the overall changes of the attention during fine-tuning, we randomly selected 1000 paragraphs from each class, averaged the cosine similarity of flattened attentions between pre-trained and fine-tuned BERT.

**Results:** Figure 5 showed the heatmap of changes for different attention heads in different layers. It showed the top three layers underwent considerable changes compared between the pre-trained and the fine-tuned BERT model. Head 10-6 (Attention Layer 10, Attention Head 6) had the largest average change. We used BertViz (Vig, 2019) to visualize Head 10-6 with an example sentence from the "Absorption" class: "*Desmopressin acetate is absorbed through the nasal mucosa.*" and summed the attention weights between the subwords for clear visualization. We noticed that a large amount of pre-trained BERT's attention focused on the period and deliminator token [SEP], which was argued as a sort of "no-op" in the model when the attention head's function is not applicable (Clark et al., 2019). After fine-tuning, the attentions switched their focus to tokens with strong task-specific semantics in absorption, such as "absorbed", "nasal", "mucosa", and so on. Similar findings were observed for other classes of "Distribution", "Metabolism", and "Excretion".

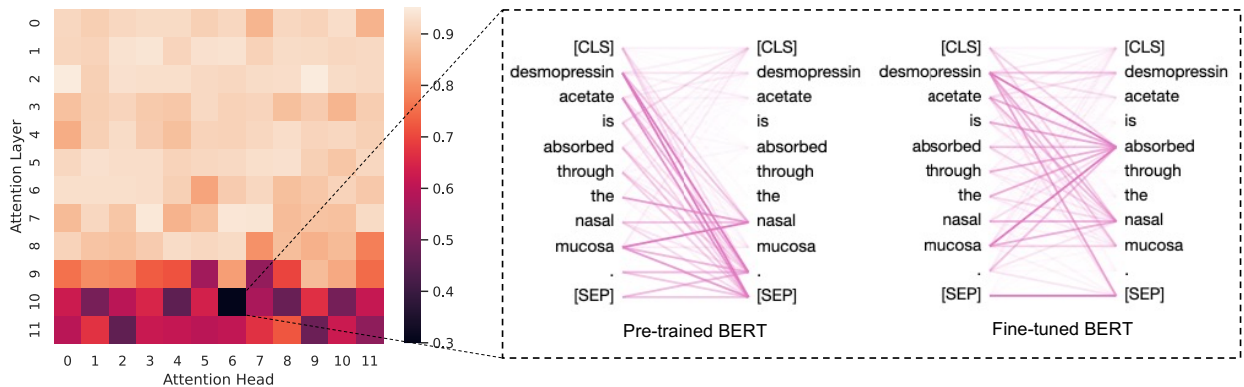

Figure 5: Per-head cosine similarity of attentions between the pre-trained and fine-tuned BERT. Head 10-6 had the largest average change, which switched their focus to tokens with strong task-specific semantics. The darkness of a line indicates the strength of the attention weight.

**Ablation Study:** Layer-based ablations, such as partial freezing, can measure the effect of the individual BERT layers on the downstream task performance. The idea of partial freezing is to keep some bottom layers fixed and only fine-tune the top layers of the model. We used partial freezing to measure how many layers in BERT were needed to fine-tune for the ADME semantic labeling task. We fine-tuned the top N layers of the BERT and preserved the weight transferred from the pretrained model in the bottom layers.

**Results:** The result shown in Figure 6 (also Table 6 for quantitative comparison) suggested that if we only fine-tuned the top 3 layers, the F1 reached 0.9, which was approaching the performance of BERT-Full (i.e., fine-tuning all the layers). It is evident that the top 3 layers contributed most to the performance improvement during the fine-tuning process (Figure 6, right). This result is consistent with the finding from the cosine similarity of attention shown in Figure 5, that attention had substantial changes in the top 3 layers.

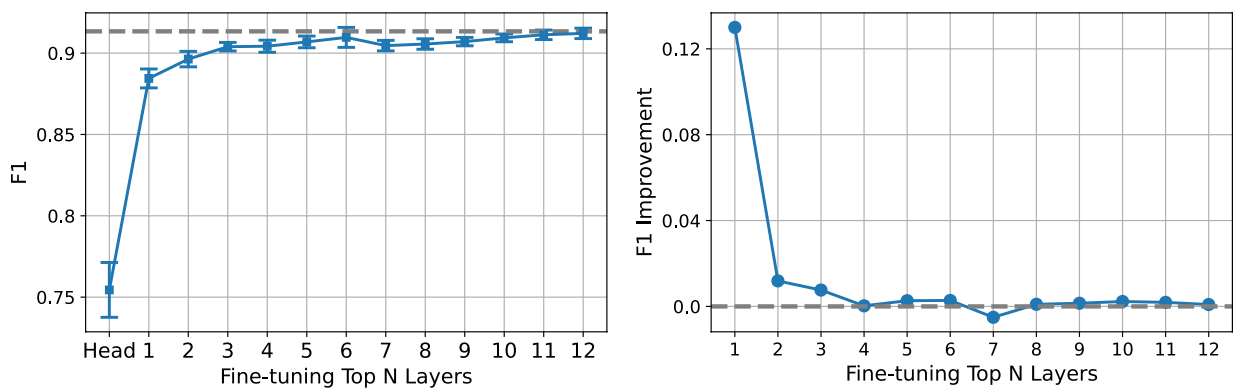

Figure 6: Performance of only fine-tuning top N layers' weights (partial freezing). The "Head" means fine-tuning the classifier only. The grey dashed line is the performance of standard fine-tuning with pre-trained BERT weights.

Table 6: Comparison of performance on fine-tuning partial layers

| Fine-tune Top N Layers | Precision | Recall | F1 |
|---|---|---|---|
| Head (Fine-tune classifier only) | 0.7811 | 0.7363 | 0.7545 |
| Top 1 Layer | 0.8839 | 0.8858 | 0.8844 |
| Top 2 Layers | 0.8971 | 0.8962 | 0.8963 |
| Top 3 Layers | 0.9043 | 0.9042 | 0.9040 |
| BERT-Full | 0.9118 | 0.9155 | 0.9134 |

### 4.3 Impact of Pre-trained BERT

To exam how the pre-trained BERT weights contribute to the overall performance in the ADME semantic labeling task, we compared three configurations of weight initialization: continuous uniform distribution $\mathcal{U}(-0.1,\ 0.1)$, original BERT initialization truncated normal distribution $\mathcal{N}(0,\ 0.02^2)$, and pre-trained BERT weights. Furthermore, we performed a similar ablation study as we did in Section 4.2, by re-initializing the top N layers with the truncated normal distribution $\mathcal{N}(0,\ 0.02^2)$, to further assess the layer-wise impact of pre-trained BERT. We compared the average performance of fine-tuning re-initialized top N layers with the standard fine-tuning with pre-trained BERT.

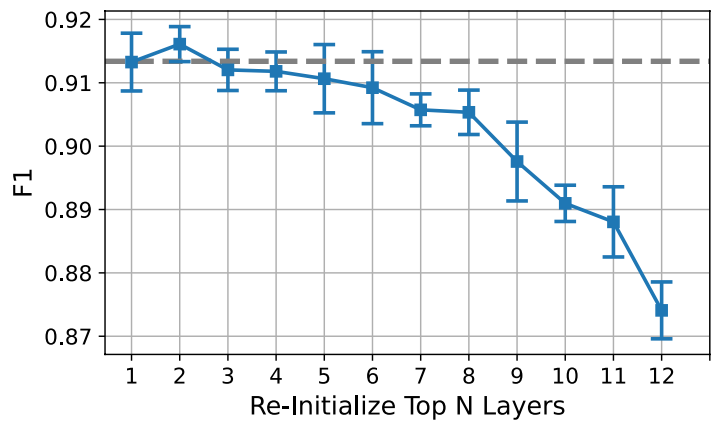

Figure 7: Performance of re-initializing top N layers' weights. The grey dashed line is the performance of standard fine-tuning with pre-trained BERT weights.

**Results:** From Table 5, we observed that the overall performance with different initialization methods varies after fine-tuning. BERT initialized with pre-trained BERT weights outperformed the models initialized with uniform or truncated normal distribution in the ADME semantic labeling task. It suggested that pre-trained BERT contains linguistic knowledge that helps solve the ADME semantic labeling task. Figure 7 showed the effect of re-initializing the top N layers with truncated normal distribution. Re-initializing the top 2 layers improved the performance, which showed that the top 2 layers are not beneficial for transfer learning during pre-training. While including more layers to re-initialize, the performance started to decrease, which showed that pre-trained weights in the middle and earlier layers are helpful to the overall performance.

## 5. Conclusion and Discussion

In this work, we proposed a novel application for ADME semantic labeling in FDA drug labeling by leveraging the state-of-the-art pre-trained language models to enhance the PSG assessment. We demonstrated that fine-tuning the pre-trained BERT model can outperform both the rule-based method and the conventional machine learning techniques, achieving up to, respectively, 26.8% and 11.6% absolute improvement in F1. Importantly, we observed excellent model performance with the independent, unseen validation dataset that were randomly sampled from unlabeled drug labels and manually annotated.

We stress that the ADME semantic labeling is a rather challenging task, as manifested by the low F1-score (0.6588) with the rule-based method and the inadequacy even with conventional machine learning techniques (0.8110 for the best-performing model). These results show that there is still a large room for improvement, and underscore the task difficulty since the ADME labeling is determined by the semantic meaning of the paragraph, rather than the simple keywords etc. Of particular note is that the performance of the BERT models on the independent unseen data is lower than those obtained by the 5-fold cross-validation, indicating that the potential biases are involved. Several kinds of bias can be introduced in the process. First, the bias is induced by label assignment. Due to various writing styles of the ADME subsection titles, there is the misassignment for the ADME labels using regular expression when preparing for the training data. For example, the paragraphs titled as "Absorption and Bioavailability" or "A. Absorption" instead of "Absorption" are assigned as "Other". Similarly, there are multi-label paragraphs such as "Absorption and Distribution", and "Metabolism and Excretion" that are also labeled as "Other". In addition, the manually annotated data inevitably contain human error. To reveal the bias induced by label assignment, we performed a simple error analysis by examining the confusion matrix. It is evident in the confusion matrix (Figure 8) that the error is mainly due to the misclassification of ADME as "Other".

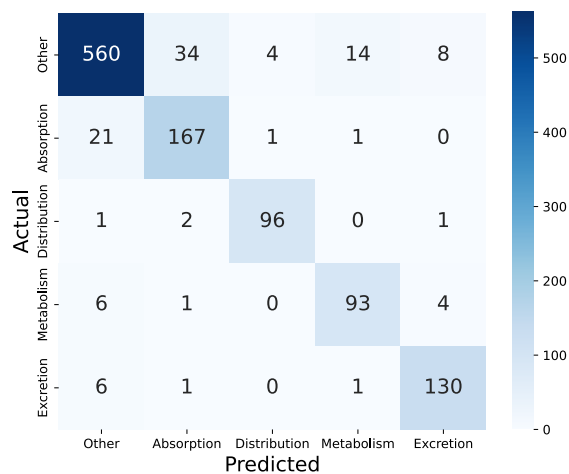

Figure 8: Confusion matrix of the BERT model on the manually annotated, unseen dataset.

Second, the bias is induced by data selection. Any resampling method for estimating error is subject to some bias. How the data are selected, particularly the data partitioning, can introduce the bias, leading to overfitting on the test set. The results can depend on a particular random choice of the data sets. Although we alleviate this issue by using 5-fold cross-validation by partitioning data into train, validation and test sets, there is still a risk of overfitting on the test set because the parameters can be tweaked until the model performs optimally. As such, knowledge about the test set can "leak" into the model and evaluation metrics no longer report on generalization performance. To measure the true performance of the model, we have performed a comprehensive evaluation with a yet-unseen validation set randomly sampled from unlabeled drug labels and manually annotated to report the results. Third, the bias is induced by data imbalance. We notice that the data is not balanced, with "Other" as the majority class. The class-imbalance poses a challenge for learning on such data as it is susceptible to an undesirable bias towards majority class, thereby leading to inferior performance for minority classes. Methods such as re-sampling and re-weighting have been used to tackle the issue, but none is effective to completely remedy the problem (Branco et al., 2015; He & Garcia, 2009; Shi et al., 2022; ValizadehAslani et al., 2022). In this work, we have adopted the stratified sampling strategy to ensure all the classes proportionally selected in order to reduce the sampling bias.

We further assessed the relative contribution of pre-training and fine-tuning to the overall performance of the BERT model in the ADME semantic labeling task using a series of analysis methods such as attention similarity and layer-based ablations. Our analysis revealed that the information learned via fine-tuning is focused on task-specific knowledge in the top layers of the BERT, whereas the benefit from the pre-trained BERT model is from the bottom layers.

When comparing different BERT models (BERT-base and BERT-large), we found that the larger model with more model parameters (BERT-large, 340 million parameters) performs better than the smaller model with fewer model parameters (BERT-base, 110 million parameters) in the ADME semantic labeling task. As such, there is a clear tradeoff between the model performance and memory usage. In addition, larger models usually take a longer time to train. For example, fine-tuning the BERT-base model on one Tesla P100-PCIE-16GB takes 7 minutes on average, whereas fine-tuning the BERT-large model takes 22 minutes, which is three times more than fine-tuning the BERT-base model. One way to speed up the training while maintaining a similar performance is to use partial freezing, which allows us to keep the bottom layers fixed and only fine-tune some top layers of the model. For our ADME task, we found that only fine-tuning the top 3 layers would suffice as they contributed the most improvement in performance during the fine-tuning process.

The observation that substantial changes in attentions were found in the top 3 layers (Figure 5) is consistent with the finding obtained from the layer ablation study. After fine-tuning, we showed that the attention was switched to the task-specific tokens in "Absorption" in the top 3 layers, which is also true for other classes including "Distribution", "Metabolism", and "Excretion". As shown in Figure 9, the attention changed its focus on tokens with strong task-specific semantics for different classes, such as "binding" in *"The plasma protein binding of pretomanid is approximately 86.4%."* from Distribution, "metabolized" in *"Lusutrombopag is primarily metabolized by CYP4 enzymes, including CYP4A11."* from Metabolism, and "excreted" and "urine" in *"Doxepin is excreted in the urine mainly in the form of glucuronide conjugates."* from "Excretion". It is worth noting that, in the attention pattern at Head 10-6 from the pre-trained BERT model for *"Doxepin is excreted in the urine mainly in the form of glucuronide conjugates."*, the token "excreted" already has strong attention weight to "urine", which is learned from common language knowledge.

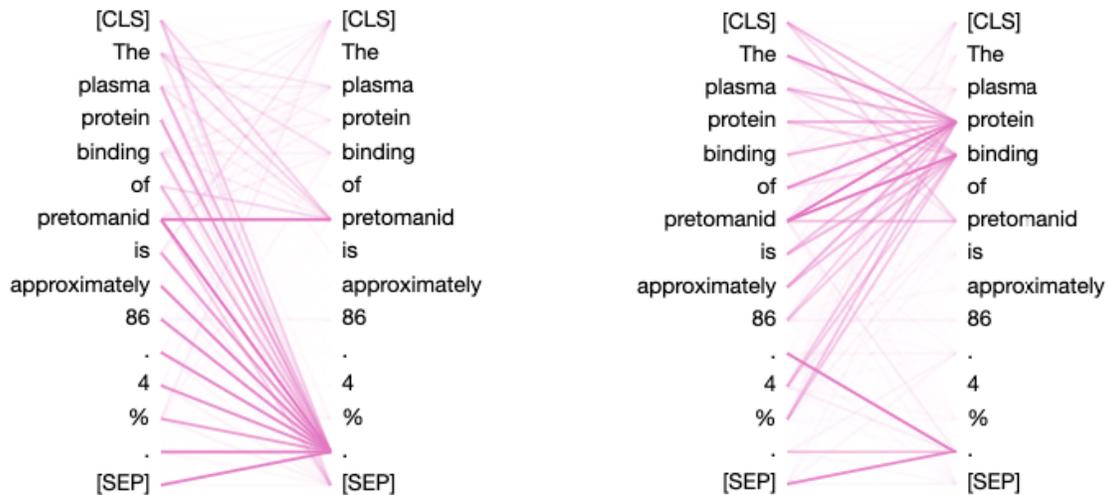

(a) Distribution

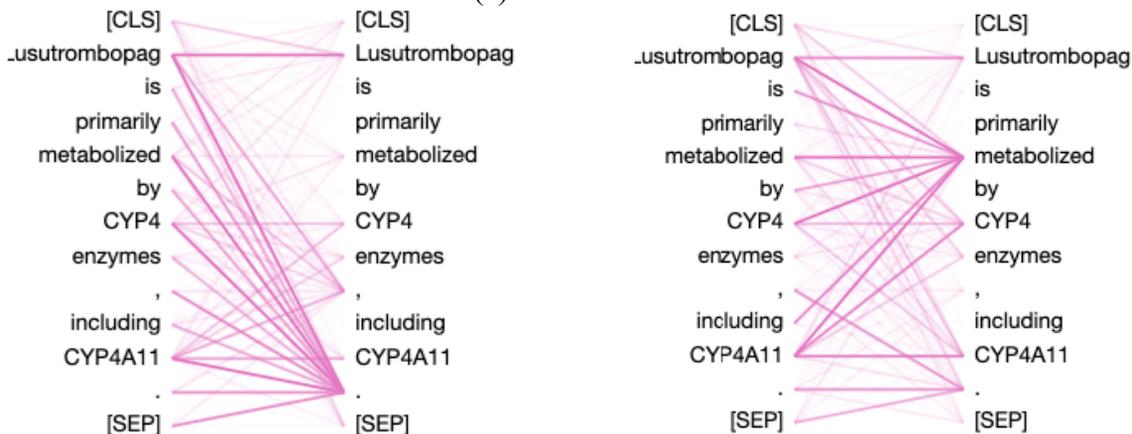

(b) Metabolism

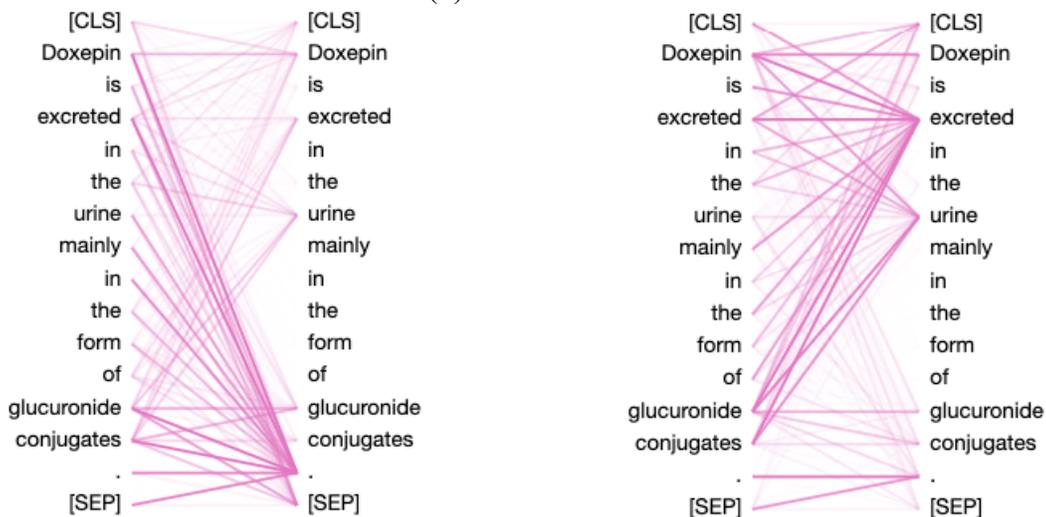

(c) Excretion

Figure 9: Comparison of attention pattern changes during fine-tuning in attention head 10-6 for three different classes (Distribution, Metabolism, and Excretion). The left pattern is from the pre-trained BERT, the right pattern is from fine-tuned BERT model. The darkness of a line indicates the strength of the attention weight.

Although the attention was switched to the task-specific tokens, mostly reflected in the top layers, the bottom layers still preserved the attention patterns transferred from the pre-trained model, such as attention to the previous/following tokens, attending broadly over many words in the sentence, and some attention heads in middle layers that correspond to linguistic phenomena (Figure 10).

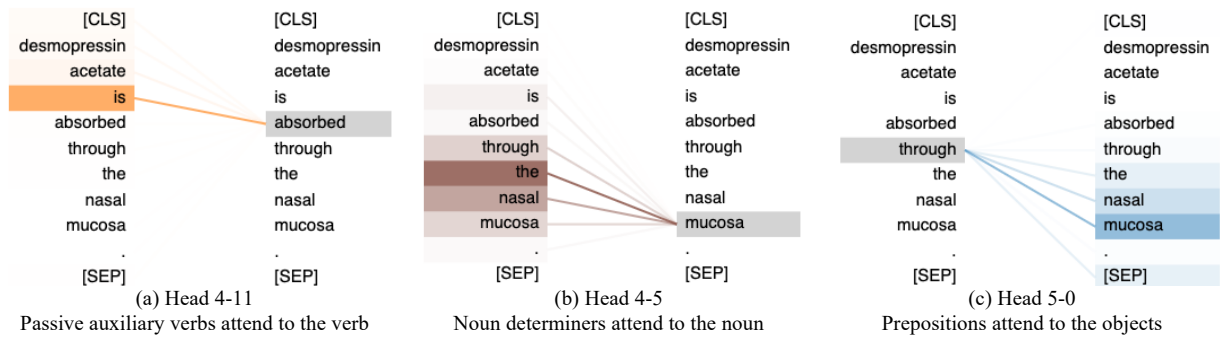

(a) Head 4-11
Passive auxiliary verbs attend to the verb

(b) Head 4-5
Noun determiners attend to the noun

(c) Head 5-0
Prepositions attend to the objects

Figure 10: BERT attention heads that correspond to linguistic phenomena transferred from the pre-trained model. The darkness of a line indicates the strength of the attention weight.

Overall, our experiments demonstrated the excellent performances of the transfer learning approach by fine-tuning the pre-trained BERT model in the ADME semantic labeling task, as compared to both the rule-based method and the traditional machine learning models. By deploying the model, we expected that the PSG assessor can retrieve the ADME information automatically from drug labeling, which can thus save a huge amount of time and effort so that the assessor can devote more time to the human intelligence required work.

**Acknowledgments**: This work was partly supported by the FDA Contract #: 75F40119C10106. The authors thank Dr. Xiajing Gong for helpful comments in the early stages of the work.

**Disclaimer:** The opinions expressed in this article are the author's own and do not reflect the view of the Food and Drug Administration, the Department of Health and Human Services, or the United States government.

# Appendix

**12.3 PHARMACOKINETICS**

*Absorption*
In pediatric patients with ALL, oral absorption of methotrexate appears to be dose dependent…

*Distribution*
After intravenous administration, the initial volume of distribution is approximately 0.18 L/kg (18% of body weight) and …

*Metabolism*
Methotrexate undergoes hepatic and intracellular metabolism to polyglutamated forms which can be converted back to methotrexate by hydrolase enzymes….

*Excretion*
Renal excretion is the primary route of elimination and is dependent upon dosage and route of administration….

FDA Application Number: NDA208400

(A) Title outside the paragraph

**12.3 PHARMACOKINETICS**

*Absorption* The estimated oral bioavailability of immediate release metoprolol is about 50% because of pre-systemic metabolism which is saturable leading to non-proportionate increase in the exposure with increased dose.

*Distribution:* Metoprolol is extensively distributed with a reported volume of distribution of 3.2 to 5.6 L/kg…

*Metabolism:* Lopressor is primarily metabolized by CYP2D6. Metoprolol is a *racemic mixture of R- and S- enantiomers,…*

*Elimination:* Elimination of Lopressor is mainly by biotransformation in the liver. The mean elimination half-life of metoprolol is 3 to 4 hours;…

FDA Application Number: NDA017963

(B) Title inside the paragraph

Figure A.1: Example of two types of ADME subsection titles can be detected by regular expression.